\pdfoutput=1

\documentclass[11pt]{article}

\usepackage[final]{acl}

\usepackage{times}
\usepackage{latexsym}

\usepackage{amsmath}
\usepackage{amssymb}
\usepackage{mathtools}
\usepackage{amsthm}
\usepackage{lipsum}  
\usepackage{enumitem}
\usepackage[capitalize,noabbrev]{cleveref}

\usepackage{algpseudocode}
\usepackage{algorithm}

\usepackage{bm}
\usepackage{xcolor}
\usepackage{multirow}
\usepackage{makecell}
\usepackage{textcomp}
\usepackage{graphicx}
\usepackage{amssymb} 
\usepackage{amsmath}
\usepackage{comment}

\usepackage{multicol}
\usepackage{newfloat}
\usepackage{caption}
\usepackage{booktabs}

\usepackage{booktabs} 
\usepackage{caption} 
\usepackage{rotating} 
\usepackage{multirow} 
\usepackage{lipsum} 
\usepackage{colortbl}
\usepackage{arydshln}

\usepackage[T1]{fontenc}

\usepackage[utf8]{inputenc}

\usepackage{microtype}

\usepackage{inconsolata}

\title{TLCR: Token-Level Continuous Reward for Fine-grained Reinforcement Learning from Human Feedback}

\author{Eunseop Yoon$^{1}$\Thanks{ Equal contribution}\hspace{17pt}  Hee Suk Yoon$^{1}$\footnotemark[1]\hspace{17pt}  SooHwan Eom$^{1}$\footnotemark[1] \\ \textbf{Gunsoo Han}$^{2}$\hspace{17pt}  \textbf{Daniel Wontae Nam}$^{2}$\hspace{17pt}  \textbf{Daejin Jo}$^{2,3}$\hspace{17pt}  \textbf{Kyoung-Woon On}$^{2}$ \\
\bf{Mark Hasegawa-Johnson}$^{4}$\hspace{17pt} Sungwoong Kim$^{3}$\footnotemark[2]\hspace{17pt}  \bf{Chang D. Yoo}$^{1}$\Thanks{ Co-corresponding authors} \\
         $^{1}$Korea Advanced Institute of Science and Technology (KAIST) \\
         $^{2}$Kakao Brain \hspace{10pt}  $^{3}$Korea University \hspace{10pt} $^{4}$University of Illinois at Urbana-Champaign (UIUC)\\
         \texttt{\{esyoon97, hskyoon, sean1105, cd\_yoo\}@kaist.ac.kr} \\ \texttt{\{jhasegaw\}@illinois.edu} \hspace{20pt} \texttt{\{swkim01\}@korea.ac.kr}}

\begin{document}
\maketitle
\begin{abstract}

Reinforcement Learning from Human Feedback (RLHF) leverages human preference data to train language models to align more closely with human essence. 
 
These human preference data, however, are labeled at the sequence level, creating a mismatch between sequence-level preference labels and tokens, which are autoregressively generated from the language model.
Although several recent approaches have tried to provide token-level (i.e., dense) rewards for each individual token, these typically rely on predefined discrete reward values (e.g., positive: +1, negative: -1, neutral: 0), failing to account for varying degrees of preference inherent to each token.
To address this limitation, we introduce \textbf{TLCR (Token-Level Continuous Reward)} for RLHF, which incorporates a discriminator trained to distinguish positive and negative tokens, and the confidence of the discriminator is used to assign continuous rewards to each token considering the context. Extensive experiments show that TLCR leads to consistent performance improvements over previous sequence-level or token-level discrete rewards on open-ended generation benchmarks. The code is publicly accessible at \url{https://github.com/EsYoon7/RLHF-TLCR}.
%
\end{abstract}

\section{Introduction}
\label{submission}

With the emergent properties of Large Language Models (LLMs) in relation to the scale of both the model and data, there has been a notable performance improvement, particularly in various language generation tasks including summarization \cite{chang2023booookscore}, translation \cite{zhang2023prompting}, and question-answering \cite{yoon2022tTHAM, yoon2023hear}. Central to these rapid performance gains in language generation is the Reinforcement Learning from Human Feedback (RLHF), which employs reinforcement learning techniques to fine-tune LLMs in alignment with human preferences \cite{rlhf, hh-rlhf, instructgpt}. 
\begin{figure}[t]
\vskip 0.2in
\begin{center}
\centerline{\includegraphics[width=0.9\linewidth]{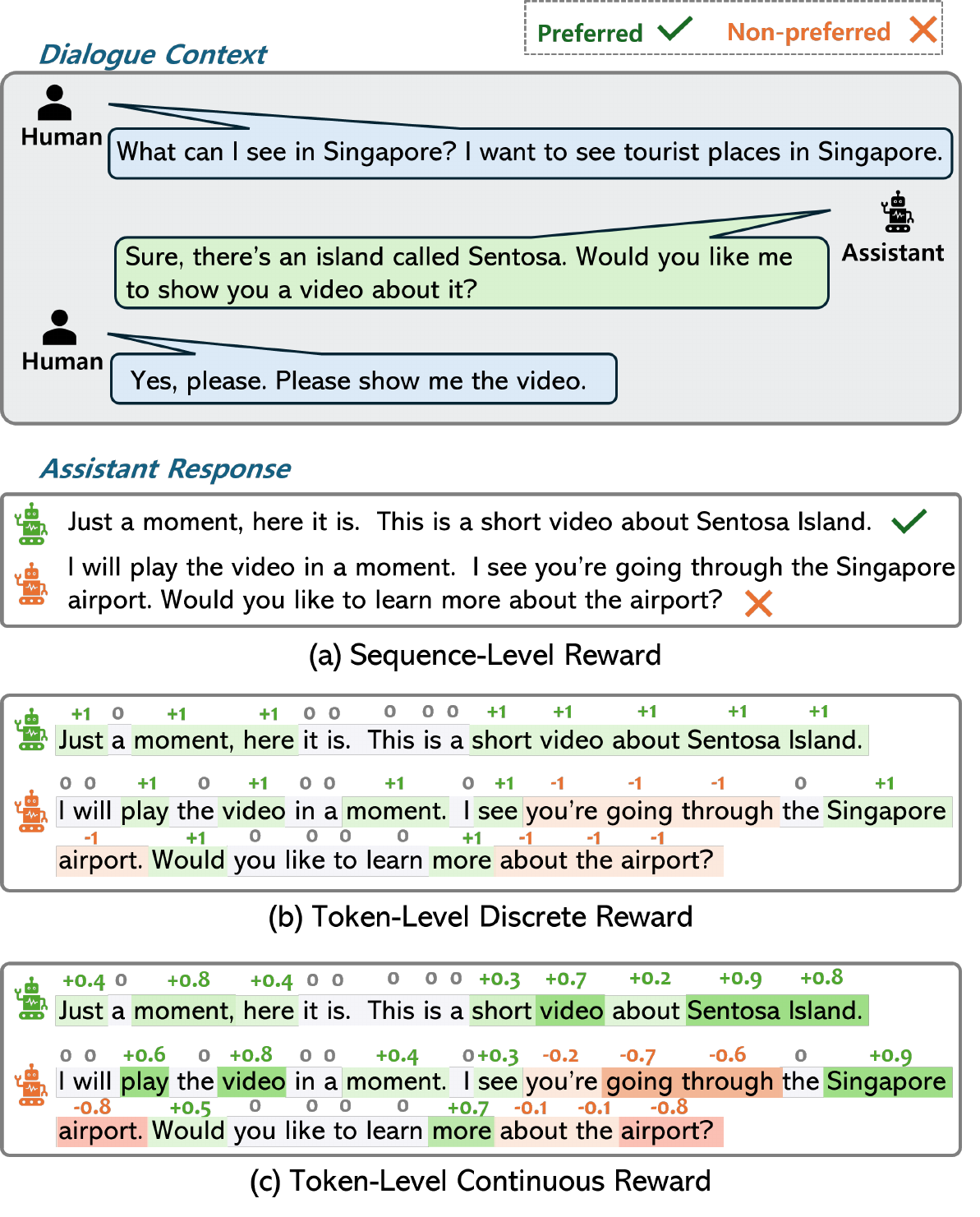}}
\caption{\textbf{Illustration of different granularity of rewards in RLHF.} (a) Sequence-Level Reward provides a singular preference value for the entire sequence. (b) Token-Level \textit{Discrete} Reward allocates fixed discrete reward values for each token. (c) Our proposed Token-Level \textit{Continuous} Reward assigns each token a continuous range of rewards.}
\label{fig:intro}
\end{center}
\vskip -0.2in
\end{figure}

However, a fundamental limitation of the current RLHF approach is its dependency on sequence-wise (i.e., sparse and holistic) rewards, which stems from the inherently holistic nature of human-labeled preference data (Figure \ref{fig:intro}-(a)). While this approach effectively directs overall model behavior, it tends to miss the detailed significance and context of specific words and phrases within sequences. This oversight suggests significant room for enhancement in existing frameworks by introducing token-wise (i.e., dense) feedback, which could more accurately capture the subtleties of language generation.

There have been several attempts to incorporate such granular feedback into the training process. Recent methods \cite{figa, drlc} employ stronger LLMs to identify regions within a model's output that require refinement, thereby enabling a denser reward signal. However, these methods still have a limitation in reflecting fine-grained preferences due to the use of a few predefined discrete reward values (Figure \ref{fig:intro}-(b)). 

To this end, we propose \textbf{TLCR (Token-Level Continuous Reward}), a novel reward model to provide token-level continuous rewards for fine-grained RLHF (Figure \ref{fig:intro}-(c)). In particular, we obtain continuous rewards efficiently and effectively from the predictive confidence of categorical preferences (positive, neutral, or negative) for each token, which are predicted by the token-level preference discriminator.
In order to train such a discriminator, our method compensates for the lack of token-level preference labels in an existing sequence-level human preference dataset by utilizing an external mature LLM, such as GPT-4 \cite{gpt4}. This process involves instructing the external LLM to revise a given output text, subsequently followed by analyzing the (1) added, (2) deleted, and (3) substituted tokens between the original and revised outputs and labeling token-level preferences.

After training the discriminator with the created token-level preference labels, we use it during reinforcement learning to provide fine-grained dense rewards to a generated text from the language model. During this process, we find that instead of using fixed reward values for each token based on its preference (e.g., using +1 for positive, -1 for negative, and 0 for neutral), assigning values based on the prediction confidence of the discriminator allows superior guidance.\textit{ Such an approach allows for a continuous range of rewards that reflect the varying degrees of preference associated with each token based on the context. }

\textbf{In detail, our contributions can be summarized as follows:}
\begin{itemize}[left=0em]
    \setlength\itemsep{0em}
    \item To address the challenges of sequence-level rewards and the inability of token-level discrete rewards to reflect varying degrees of token preference, we introduce Token-Level Continuous Reward (TLCR), that provides a continuous spectrum of rewards for each token during Reinforcement Learning from Human Feedback (RLHF).
    \item For this, TLCR leverages confidence values from a token-level preference discriminator, and to train the discriminator, we use an external mature language model to refine responses, then assign token-wise preference labels by calculating the minimum edit distance between the original and revised responses.
    \item Through extensive experiments and analysis, we show the effectiveness of TLCR in enhancing fine-grained RLHF compared to the traditional sequence-level and token-level discrete reward mechanisms. \textit{To the best of our knowledge, our work is the first to achieve substantial performance gain on open-ended generation benchmarks using token-level rewards.}
\end{itemize}

\section{Related Works}
\subsection{Reinforcement Learning from Human Feedback.}

Foundation models have achieved notable performance improvement due to increased model size and data volume. With these advancements, ensuring safe and reliable real-world applications has become a priority \cite{yoon2024ctpt}. In this aspect, Reinforcement Learning from Human Feedback (RLHF) aims to integrate human judgments and preferences into the reinforcement learning framework to train models, such as language models, to align more closely with human values and expectations \cite{rlhf, hh-rlhf, instructgpt}. In this approach, a reward model is initially trained based on human feedback, usually pairwise preference ranking, using the Bradley-Terry preference model \cite{btmodel}. This reward model is then utilized to guide the model via reinforcement learning frameworks, usually Proximal Policy Optimization (PPO) \cite{ppo}.

However, despite its effectiveness, RLHF has complexities involving multiple stages and models, demanding substantial resources.
These challenges are pronounced in large-scale implementations, with recent research highlighting potential instability issues \cite{secrets_of_rlhf1, rm_overoptimization}.
To tackle such limitations, Hydra-PPO \cite{hydra-ppo} utilizes switchable architecture to integrate multiple models involved in PPO training into a single LLM, while ReMax \cite{remax} adapts REINFORCE \cite{reinforce1}, simplifying the framework by eliminating the need for extra models used in PPO.

Recent studies have also been exploring alternatives to conventional reinforcement learning (RL) frameworks for efficient training.
Direct Preference Optimization (DPO) \cite{dpo} simplifies LLM training by using log-likelihood as an implicit reward, avoiding the need for separate reward modeling.

\subsection{Fine-grained Feedback}


The majority of LLM fine-tuning with human feedback, including RLHF \cite{instructgpt} and DPO \cite{dpo}, typically utilize trajectory-wise rewards learned from pairwise preference learning. This approach involves assessing and providing feedback on LLM responses as a whole rather than each individual word or phrase within the sequence.

While trajectory-wise rewards have been effective in guiding the overall behavior of LLMs, they may present limitations in terms of the granularity and specificity of the feedback. This might lead to less precise tuning of the model, since specific aspects of responses that could be improved or are particularly well-aligned with desired outcomes may not be individually identified and adjusted. This highlights the potential need for more fine-grained feedback mechanisms in RLHF for LLMs.
For instance, \citet{finegrained} shows that fine-grained rewards enable more detailed and specific guidance for each phrase of the LLM response, potentially leading to more nuanced and effective learning and alignment with human preferences and intentions.

One of the most significant challenges in utilizing fine-grained rewards for aligning LLMs is the lack of datasets with fine-grained feedback, since most existing datasets provide trajectory-wise or holistic labels. This gap in data availability necessitates innovative approaches to generate or infer fine-grained signals from the available trajectory-wise human feedback.
\citet{preference-grounded} defines the trajectory-wise reward as an aggregation of individual token-wise rewards, which are learned through standard preference-based reward model training. However, selecting an optimal aggregation function for token-wise rewards is not straightforward and depends largely on heuristic approaches. Also, the lack of additional fine-grained information or constraints will result in a sub-optimal reward function.

To address this issue, recent techniques leverage external advanced LLMs, such as GPT-4 \cite{gpt4}, to provide more detailed and accurate fine-grained insights for learning dense rewards. Approaches such as FIGA \cite{figa} and RLMEC \cite{rlmec} utilize off-the-shelf LLMs to revise the generated response with minimal edits, pinpointing areas within responses that need fine-tuning and providing specific guidance at the token level. Meanwhile, DRLC \cite{drlc} asks external LLMs to identify the positive and negative segments within an original response, utilizing them as intrinsic rewards to guide model adjustments.
On the other hand, ABC \cite{abc} extracts fine-grained credits from the trained reward model itself, particularly using attention maps for token-specific reward reallocation.

Despite the advancements these methods have limitations. Some can only  offer provide predefined values for rewards (such as FIGA, and DRLC) and cannot clearly differentiate between positive and negative impacts (such as ABC). Others suffer from an inadequate reward model, necessitating a heavy dependency on stronger regularization (such as RLMEC).
In contrast, our approach offers continuous-scale rewards that encompass both positive and negative feedback without the need for excessive regularization, leading to a more refined and effective learning.

\section{Preliminary}
In this work, we focus on aligning the language model to human preference in terms of helpfulness and harmfulness of the generated texts through Reinforcement Learning from Human Feedback (RLHF). 
\begin{figure*}[ht!]

\centerline{\includegraphics[width=0.9\linewidth]{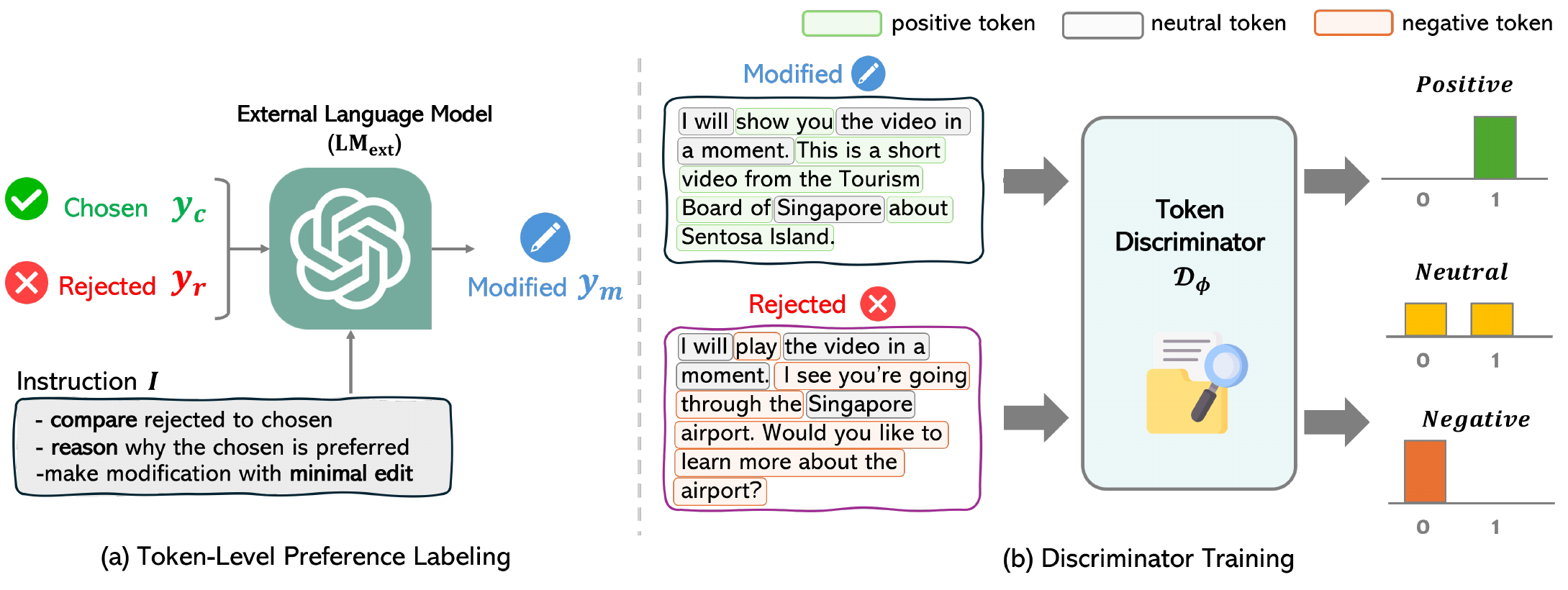}} 
\caption{\textbf{Illustration of the training procedure of the discriminator used in TLCR (Token-Level Continuous Reward).} (a) Using the sequence-level labeled dataset, we utilize an external mature language model $\text{LLM}_\text{ext}$ as a reviser to obtain token-level preference labels. $\text{LLM}_\text{ext}$ is instructed to compare the chosen ($y_c$) and rejected response ($y_m$), reason why the chosen is preferred, and create modified response $y_m$ by modifying the rejected response with minimal editing. Using the Levenshtein Distance between $y_r$ and $y_m$, we assign token-wise preference labels based on whether the tokens have been added, deleted, or substituted. (b) With the token-wise preference label created from the previous step, we train a discriminator to discriminate positive, neutral, and negative tokens.}
\label{fig:method1}
\end{figure*}
\paragraph{Language Generation as an MDP} We consider the language generation procedure as a Markov Decision Process (MDP) as detailed in \citet{IsRL}, which is defined by the tuple $(\mathcal{S}, \mathcal{A}, \mathcal{P}, \mathcal{R}, \gamma, \mathcal{T}_\text{max})$ using finite predefined vocabulary tokens $\mathcal{V}$. An episode in the MDP starts with a human instruction or a history of human-assistant conversation pairs followed by a human instruction $x=(x_0,...,x_l)$. This is used as our initial state $s_0 = (x_0,...,x_l)$, where $s_0 \in \mathcal{S}$ and $\mathcal{S}$ represents every possible states with $x_i \in \mathcal{V}$. At time step $t$, an action in the environment $a_t \in \mathcal{A}$ is generated by the next token prediction given current state $s_t$ by a policy Language Model (LM) $\pi_{\theta}$ from $\mathcal{V}$. For $t \geq 1$, the transition function $\mathcal{P}: \mathcal{S} \times \mathcal{A} \mapsto \Delta \mathcal{S}$ appends $a_t$ to the end of the state $s_{t} = (x_0,...x_l, a_0,...,a_{t-1})$. This process continues until the current $t$ exceeds $\mathcal{T}_\text{max}$ or the end of sentence (<eos>) token is generated, yielding the generated sequence $y=(a_0,...,a_T)$. A reward function $\mathcal{R}: \mathcal{S} \times \mathcal{A} \mapsto \mathbb{R}$ assigns a value to each transition. 

\paragraph{Sequence-level Reward in RLHF} Typically, in RLHF for language models, a sequence-level (i.e., sparse) reward model is used, which maps the concatenation of the input $x$ and the generated output $y$ 
to a single scalar value. This scalar value is solely assigned to the last generated token or the end of sentence (<eos>) token $a_T$. Formally, this process of sparse reward modeling used in previous RLHF frameworks can be defined as: 
\begin{align}\label{eq:holistic_r}
    r_t= 
\begin{cases}
    R_{\phi}(x, y),& \text{if } t = T\\
    0,              & \text{otherwise.}
\end{cases}
\end{align} 

\paragraph{Token-Level \textit{Discrete} Reward in RLHF} In contrast to the sparse reward approach, token-level discrete rewards in RLHF assign a predefined discrete value at each time step $t \in [0,T]$, reflecting the preference for each token, formalized as:
\begin{align}\label{eq:token_r}
r_t=
\begin{cases}
+1, & \text{if $a_t$ is positive}\\
-1, & \text{if $a_t$ is negative}\\
0, & \text{if $a_t$ is neutral.}
\end{cases}
\end{align}

\section{TLCR: Token-Level \textit{Continuous} Reward}

In this section, we introduce an RLHF framework that utilizes a discriminator-based token-level reward model that allows continuous-scale dense reward signals for RLHF. Section \ref{method1} details the procedure for obtaining token-level preference labels. In Section \ref{method2}, we elaborate on the training procedure of the token-level preference discriminator. Subsequently, Section \ref{method3} outlines how the trained discriminator is utilized for continuous dense reward signals during the RL phase.

\subsection{Token-Level Preference Labeling} 
\label{method1}
Human preference data \cite{hh-rlhf, ultrafeedback} is generally comprised of a prompt $x = (x_0,...,x_l)$, accompanied by its chosen response $y_c=(a_0^c,...a_{T_c}^c)$ and the rejected response $y_r=(a_0^r,...,a_{T_r}^r)$. The categorization of responses as being labeled either "chosen" or "rejected" reflects the holistic evaluation of preferences.

As illustrated in Figure \ref{fig:method1}-(a), we utilize an external mature Language Model $\text{LM}_\text{ext}$ as a reviser to assign token-wise preference labels for the response. The process begins by presenting the model with both the prompt and the rejected response to refine the rejected response to align more closely with human preferences. Considering the difficulty in explicitly defining "good human preference," we reduce the ambiguity during the process by also providing the chosen response $y_c$ to $\text{LM}_\text{ext}$. This allows the revising model to compare the rejected and chosen responses, understand the reasons behind the chosen response's preference, and adjust the rejected response to better meet human preference criteria. Throughout this refinement process, we instruct the model to make minimal modifications to the rejected response. This minimal modification ensures the core message remains intact while optimizing preference alignment, making it easy to identify which particular tokens are responsible for the output preference, a more straightforward approach than comparing $y_c$ and $y_r$ directly. Formally, given an instruction $I$ (the full details of the instruction used are described in Appendix \ref{sec:appendix}), prompt $x$, rejected response $y_r$, and chosen response $y_c$, the $\text{LM}_\text{ext}$ produces the modified response given as
\begin{equation}
    y_m = (a_0^m,...,a_{T_m}^m) = \text{LM}_\text{ext}(I, x, y_r, y_c).
\end{equation}

Motivated by previous works \cite{levenshtein, figa}, once we obtain the modified response $y_m$, we use the Levenshtein Distance ($L_D$) \cite{levenshtein1965binary} to determine the preference of each token. The $L_D$ calculates the minimum number of edits—(1) additions, (2) deletions, or (3) substitutions—required to transform one sentence into another through dynamic programming. By identifying the edited tokens regarding $L_D$, we can determine which tokens contribute to a response being preferred or rejected.

When calculating the $L_D$ of going from $y_r$ to $y_m$, we assign a token-wise preference labels $p_r = (p_0^r,...p_{T_r}^r)$ and $p_m=(p_0^m,...,p_{T_m}^m)$ for $y_r$ and $y_m$, respectively. Specifically, 
\begin{align}
\small
p_t^r= 
\begin{cases}
    \text{negative},& \text{if } a_t^r \text{ is \textbf{deleted} or \textbf{substituted}}\\
    \text{neutral},              & \text{otherwise,}
\end{cases}
\end{align}
\begin{align}
\small
p_t^m= 
\begin{cases}
    \text{positive},& \text{if } a_t^m \text{ is \textbf{added} or \textbf{substituted}}\\
    \text{neutral},              & \text{otherwise.}
\end{cases}
\end{align}

\subsection{Discriminator Training}
\label{method2}
Following the token-level preference labeling process, we train a discriminator \textbf{$D_{\phi}$} that differentiates between these positive and negative tokens (Figure \ref{fig:method1}-(b)). The output space of \textbf{$D_{\phi}$} is binary, where the prediction of 1 indicates positive and 0 indicates negative preference of the token.

To train the discriminator, we employ a binary cross-entropy loss function. Formally, for a given token $a_t$ with its corresponding preference label $p_t$, the loss function $\mathcal{L}$ for \textbf{$D_{\phi}$} is defined as follows:
\begin{equation}
\begin{split}
\mathcal{L} &= -p_t \log(D_{\phi}(a_t|x, a_{0:t-1})) \\
            &\quad -(1 - p_t) \log(1 - D_{\phi}(a_t|x, a_{0:t-1})),
\end{split}
\end{equation}
where $D_{\phi}(a_t|x, a_{1:t-1})$ represents the discriminator's predicted probability that the $t$-th token $a_t$ is positive given the previous contexts, and \textit{$p_t = 1$ for positive labels}, \textit{$p_t = 0$ for negative labels}.

For tokens labeled as neutral, we introduce a soft labeling strategy to encourage the discriminator to remain unbiased towards either class. Specifically, neutral tokens are assigned a uniform probability distribution (i.e., \textit{$p_t=0.5$ for neutral labels}), implying an equal likelihood of being positive or negative. 
\begin{figure}[t!]
\begin{center}
\centerline{\includegraphics[width=1.0\linewidth]{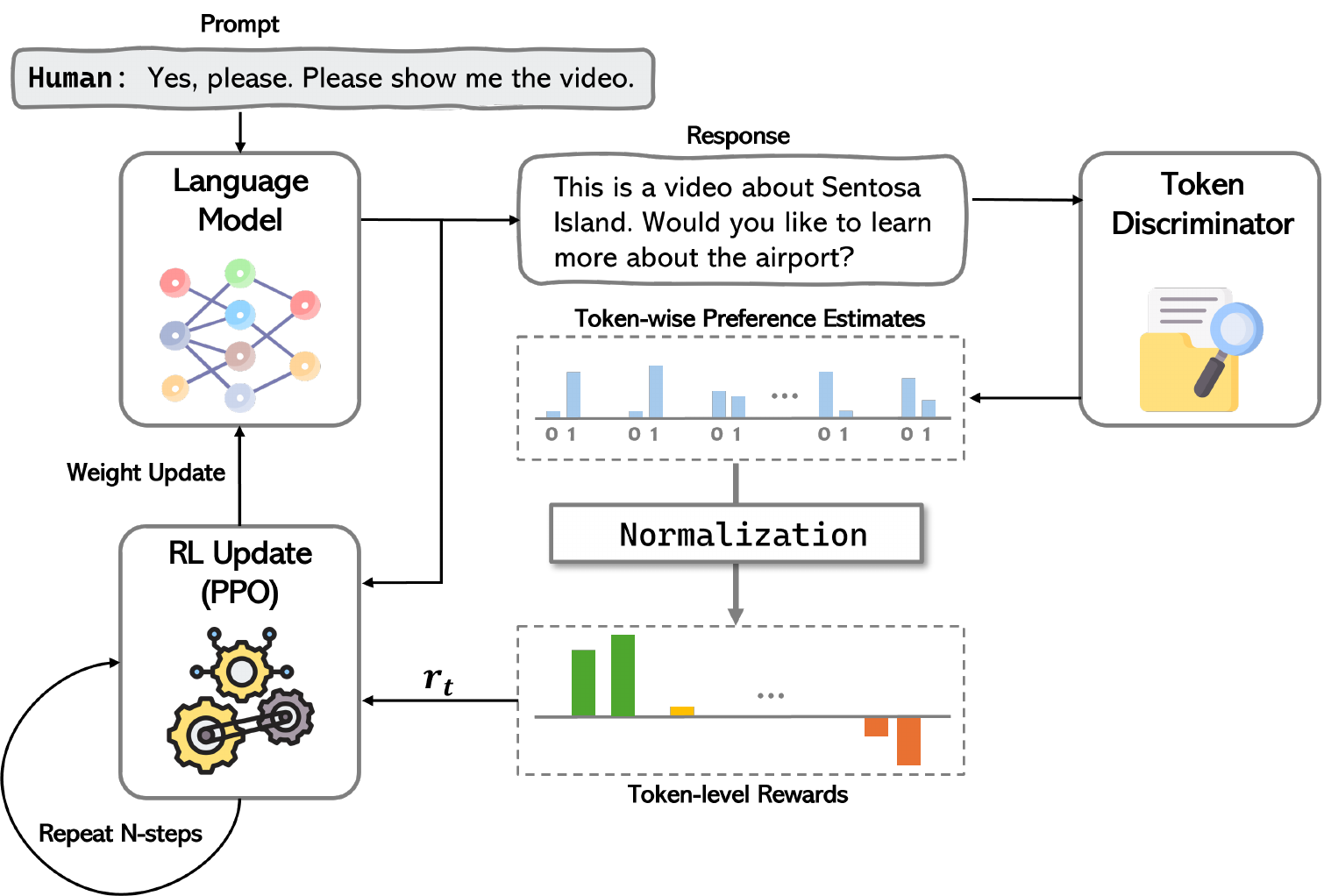}}
\caption{
\textbf{Illustration of using the discriminator for assigning token-level continuous reward during PPO.} The discriminator's prediction probability of a token being positive undergoes normalization to fit a scale from -1 to 1. A value near -1 signifies an unfavorable preference, near 1 suggests a favorable preference, and around 0 denotes a neutral preference.
}
\label{fig: step3}
\end{center}
\end{figure}

\subsection{Discriminator-Guided Token-Level Continuous Reward}
\label{method3}
As illustrated in Figure \ref{fig: step3}, we utilize the trained discriminator to obtain token-level signals during the reinforcement learning phase.
In detail, our discriminator-guided token-level reward mechanism employs a normalization formula,
\begin{equation}
r_t = 2 \cdot D_{\phi}(a_t|x, a_{0:t-1}) - 1 \quad \forall t \in [0, T],
\end{equation}
to translate the discriminator's confidence in token preference classification into a scale ranging from -1 to 1.
A maximum reward of 1 is granted to tokens identified with high confidence as aligning positively, signaling their strong adherence to encouraging the model to favor such tokens in future responses. Conversely, tokens considered negative, marked by a confidence score of 0, receive the minimum reward of -1, indicating a deviation from the desired preferences and advising the model to avoid such tokens. Tokens that the discriminator views with neutral confidence scored at 0.5 are assigned a reward of 0, reflecting their ambiguous contribution to preference alignment and suggesting no immediate need for the model to alter its approach for such tokens.


Given our token-wise reward approach, we adapt our reinforcement learning phase to align with the original PPO \cite{ppo} implementation, diverging from the conventional RLHF framework with a sparse reward setting (contextual bandit) \cite{instructgpt}.


\section{Experimental Settings}
This section presents the datasets, models, compared baselines, training details, and the evaluation metrics used in the paper.
\paragraph{Dataset} Throughout the experiments, we use the \textit{full-hh-rlhf} dataset \cite{hh-rlhf} which is centered on improving the helpfulness and harmlessness of the language model generation. This dataset includes 112k training instances and 12.5k instances for evaluation. Every instance in the dataset features a prompt, along with a chosen response considered preferable and a rejected response, offering a clear basis for performing model alignment for better helpfulness and harmlessness. Following the setting in \cite{remax}, we randomly divide the dataset into three parts: 20\% for supervised fine-tuning (SFT), 40\% for reward model learning, and 40\% for reinforcement learning for reward maximization.

\paragraph{Model} During the Supervised Fine-Tuning (SFT) phase we use \textit{Llama-2-7B} \cite{llama2} as our base model. Similarly, for the reward model training, we initialize our discriminator $D_{\phi}$ with \textit{Llama-2-7B}. During the RL phase, we initialize our policy model with the trained SFT model. For the external language model $\text{LM}_\text{ext}$, we use \textit{gpt-4-0125-preview}\footnote{\href{https://platform.openai.com/docs/models/gpt-4-and-gpt-4-turbo}{https://platform.openai.com/docs/models/gpt-4-and-gpt-4-turbo}}. 

\paragraph{Training Details}
We use DeepSpeed-Chat \cite{deepspeed} framework for performing RLHF, where PPO is the default algorithm. Within this framework, our proposed TLCR is integrated to provide token-level rewards to enhance the PPO algorithm. Experimental details are presented in Appendix \ref{appendix: experimental_detail}.

All experiments were conducted 3 times with different random seeds, and the average value is reported. The experiments were conducted using 8 x NVIDIA A100 80GB PCIe.

\paragraph{Baselines} 
For comparison, we evaluate against a variety of baselines: \textit{SFT} and \textit{DPO} \cite{dpo}, where no preference data is utilized; sequence-level reward PPO (\textit{PPO$_\text{seq}$}) \cite{ppo} and \textit{ReMax} \cite{remax}, both employing sparse rewards; as well as \textit{FIGA} \cite{figa}, which introduce fixed token-level rewards.

Moreover, to verify that the performance improvements are not solely due to the use of GPT-4 for creating synthetic data, we conducted an additional experiment, labeled \textit{PPO$_\text{synthetic}$}. This experiment used pairs of rejected samples and their synthetically modified responses generated by GPT-4, applied to traditional sequence-level PPO \cite{ppo}. Additionally, we include \textit{TLCR$_\text{fixed}$} as a variant of our approach using fixed token-level rewards. Specifically, \textit{TLCR$_\text{fixed}$} assigns a reward of +1 to the positive, -1 to the negative, and 0 to the neutral token instead of normalized confidence values. A token is considered neutral if the max confidence is below 0.6. 



\paragraph{Evaluation} Automatic evaluation of instruction-following language models are becoming a primary metric for preference evaluation with leaderboards\footnote{\href{https://huggingface.co/spaces/lmsys/chatbot-arena-leaderboard}{https://huggingface.co/spaces/lmsys/chatbot-arena-leaderboard}} for ranking. In light of this trend, , we report \textit{MT-Bench} \cite{mtbench} and \textit{AlpacaEval (alpaca\_eval\_gpt4)} \cite{alpaca_eval} for our evaluations. Also, we perform human evaluation on 100 random samples from the \textit{full-hh-rlhf} test set.



\section{Results}
\begin{table}[t]
 \setlength{\tabcolsep}{3.0pt} 
 \small
  \centering
  \begin{tabular}{l || c c | c}
    \toprule
    \multirow{1}{*}{Method} & Turn1 & Turn2 & \textbf{Overall} \\
    \Xhline{2\arrayrulewidth}
    SFT & 4.78 & 3.01 &3.90 \\
    DPO \cite{dpo}& 5.10 & 3.56 & 4.33 \\
    PPO$_\text{seq}$ \cite{ppo}  & 5.29 & 4.08 &4.68 \\
    ReMax \cite{remax}& 5.66 & 3.86 & 4.76 \\
    FIGA \cite{figa} & 5.11 & 3.57 & 4.35 \\
    PPO$_\text{synthetic}$ & 4.90 & 3.53 & 4.21 \\
    TLCR$_\text{fixed}$ & 5.47 & 4.3 & 4.89 \\
    \textbf{TLCR (ours)} & \textbf{5.71} & \textbf{4.38} & \textbf{5.04}\\
    \bottomrule
  \end{tabular}
  \caption{\textbf{Turn1, Turn2, and Overall Evaluation results on MT-Bench \cite{mtbench}.}}
  \label{tab:mtbench}
\end{table}
\subsection{Results on MT-Bench}
Table \ref{tab:mtbench} shows the comparison of the MT-Bench \cite{mtbench} evaluation results. MT-Bench is a multi-turn benchmark that measures the ability of LLMs to engage in coherent, informative, and engaging conversations. The process of evaluation involves prompting GPT-4 to act as judges and asses the quality of the models' responses on 80 questions on a scale of 1 to 10. The result shows that among the baselines, TLCR achieves the highest preference scores with the overall score being 5.04. Notably, compared to the sequence-level (i.e., sparse) rewarded PPO (PPO$_\text{seq}$), our TLCR, which is a token-level (i.e., dense) rewarded PPO, achieves 0.36 increase in performance.

Compared to the results from SFT, PPO$_\text{synthetic}$ shows better performance, yet it underperforms the PPO$_\text{seq}$ trained on the original data. Furthermore, a significant performance gap was observed between PPO$_\text{synthetic}$ and TLCR, which was trained on the same synthetic data. Moreover, our TLCR outperforms TLCR$_\text{fixed}$ by 0.15. \textit{These results indicate that the modified synthetic dataset from GPT-4 alone cannot give a satisfactory signal during the alignment process without our targeted continuous token-level reward.}


\subsection{Results on AlpacaEval}
\begin{figure}[t!]
\begin{center}
\centerline{\includegraphics[width=1.0\linewidth]{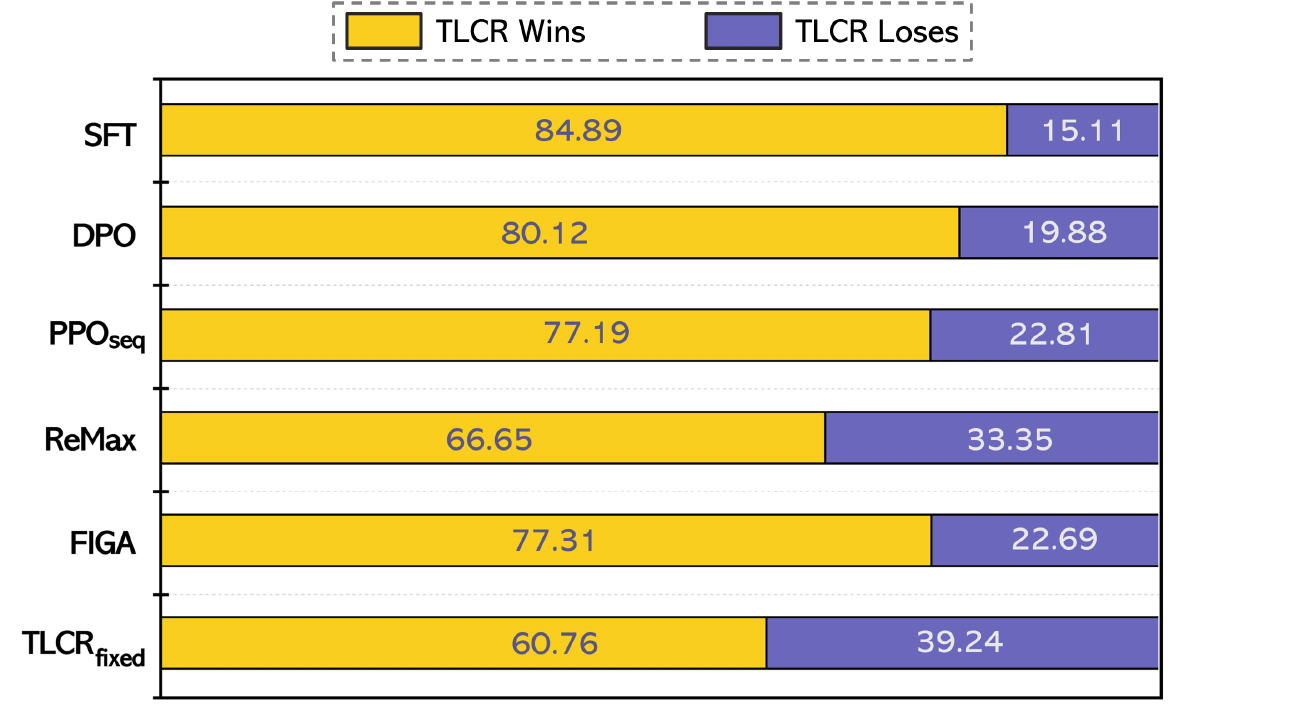}}
\caption{\textbf{Evaluation on test questions from AlpacaEval dataset \cite{alpaca_eval}.}}
\label{fig: alpaca}
\end{center}
\end{figure}

Figure \ref{fig: alpaca} shows the evaluation results for 805 test questions from the AlpacaEval dataset \cite{alpaca_eval}. This evaluation method entails presenting GPT-4 with two responses generated by different models and determining the binary win rate based on GPT-4's preference. We achieve the higher win rate among all baselines; TLCR achieves win rates of 84.89\%. 80.12\%, 77.19\%, 66.65\%, 77.31\%, and 60.76\% over SFT, DPO, PPO$_{seq}$, ReMax, FIGA, and TLCR$_\text{fixed}$ respectively.

\begin{figure}[t!]
	\centering
    	\includegraphics[width=1.0\linewidth]{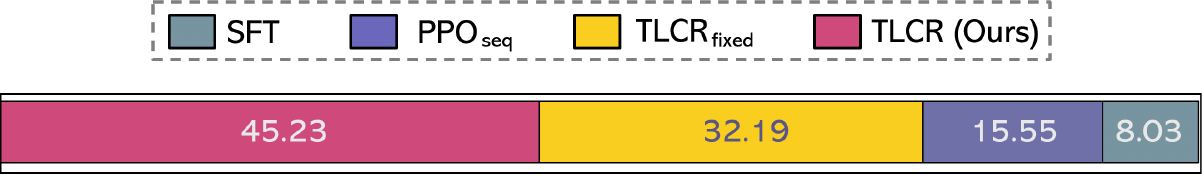}
	\caption{\textbf{Human evaluation results on 100 random samples from full-hh-rlhf test set.} Five annotators were tasked with selecting the most preferred response generated by different methods. We report the average proportion of preferences chosen to each method's outputs} 
	\label{fig: human_eval}
\end{figure}

\subsection{Human Evaluation}
We present the human evaluation result in Figure \ref{fig: human_eval}. Given the generations from SFT, $\text{PPO}_\text{seq}$, $\text{TLCR}_\text{fixed}$, and TLCR, five human annotators\footnote{Human annotators all have TOEFL iBT score above 100 and possess a bachelor's degree or higher.} are asked to choose the most preferred answer. We report the ratio of samples selected as the most preferred. The results show that TLCR achieved the highest preference rate, receiving an average of 45.23\% of the total votes.

\begin{figure}[t]
	\centering
    	\includegraphics[width=1.0\linewidth]{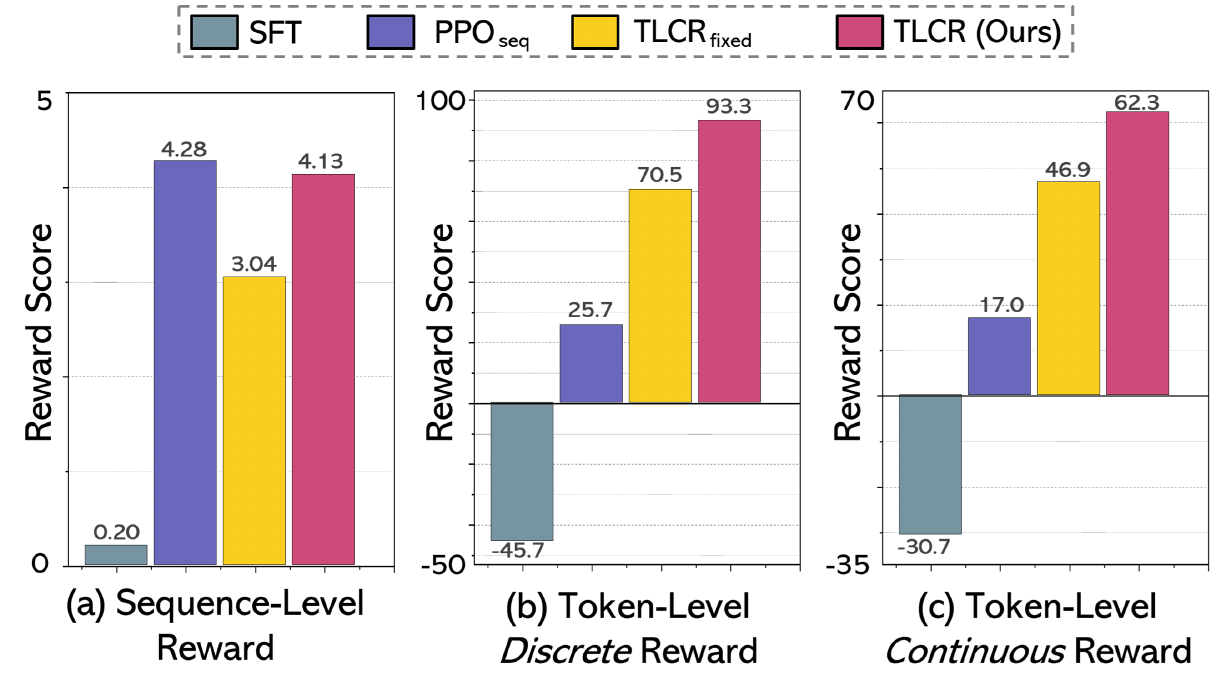}
	\caption{\textbf{Reward score evaluation comparison} using three different reward schemes: (a) Sequence-Level Reward, (b) Token-Level Discrete Reward, and (c) Token-Level Continuous Reward.} 
	\label{fig: reward comparison}
\end{figure}

\section{Ablation Studies}

\subsection{Reward Comparison} 
In Figure \ref{fig: reward comparison}, we compare reward scores across different methods: SFT, PPO, TLCR$_\text{fixed}$, and our proposed TLCR. Across all evaluated reward metrics, SFT consistently records the lowest scores due to its lack of training for maximizing any specific reward score.

Figure \ref{fig: reward comparison}-(a) shows the reward score obtained using sequence-level rewards, which were the direct maximization target for PPO$_\text{seq}$ during reinforcement learning. Notably, TLCR (ours), despite not being explicitly maximized with this reward, shows a reward score closely competitive to that of PPO$_\text{seq}$, obtaining a score of 4.13 compared to 4.28 from PPO$_\text{seq}$. On the other hand, TLCR$_\text{fixed}$ lags behind in achieving comparable sequence-level reward scores, with only obtaining a score of 3.04.

Moving to Figure \ref{fig: reward comparison}-(b), we look into the evaluation based on token-level discrete rewards, a value for which TLCR$_\text{fixed}$ explicitly maximized. Here, the limitations of sequence-level feedback become apparent, as evidenced by PPO$_\text{seq}$'s underperformance (achieving a score of only 25.7) in enhancing token-level generation quality. Notably, TLCR also exceeds this aspect, effectively capturing token-level preference.

Lastly, Figure \ref{fig: reward comparison}-(c) shows the outcomes evaluated on token-level continuous rewards, the core metric our TLCR model was trained to maximize. Consistent with previous trends, PPO$_\text{seq}$ shows insufficient results (with a score of 17.0), further emphasizing the shortcomings of sequence-level feedback in refining token-level outputs. TLCR achieves the highest score in this category as well.

Overall, these findings reinforce the inherent limitations of traditional sequence-level and token-level discrete reward feedback mechanisms and demonstrate the advantages of our TLCR approach in achieving token-level improvements.

\begin{figure}[t]
	\centering
    	\includegraphics[width=1.0\linewidth]{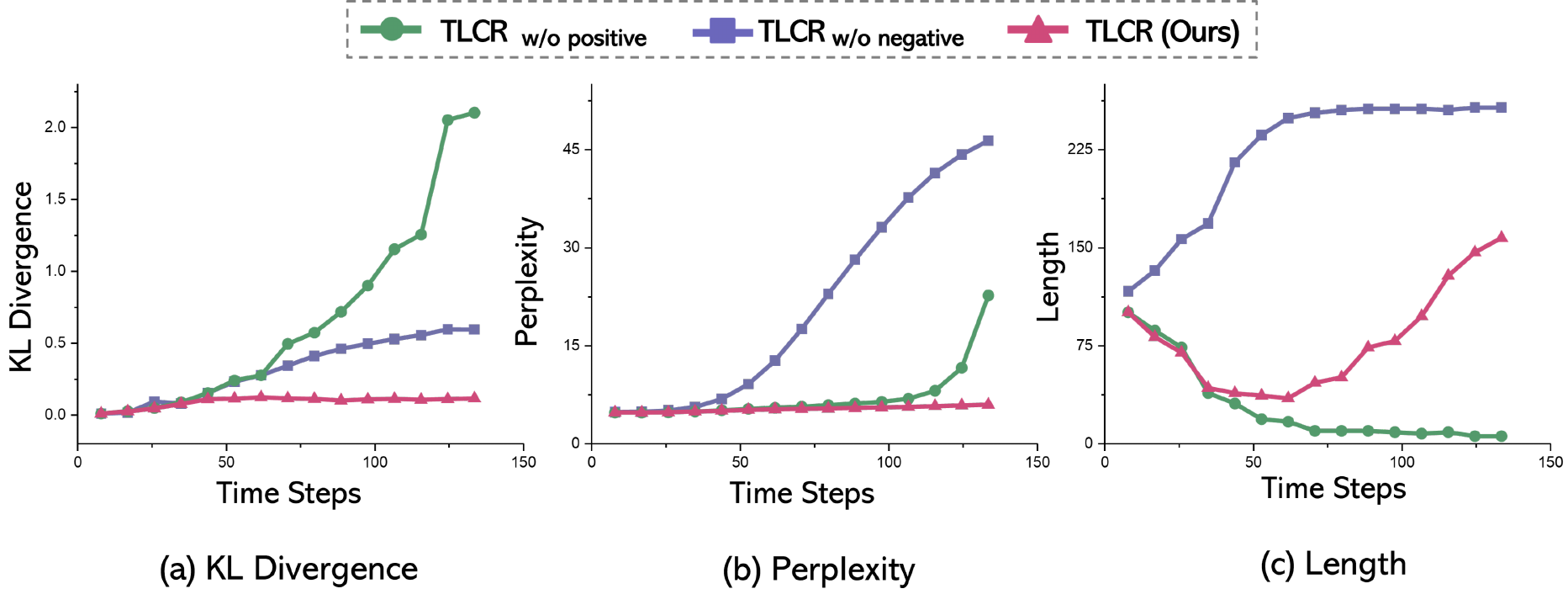}
	\caption{\textbf{PPO training curve (perplexity, length, KL divergence) for TLCR under various conditions.}} 
	\label{fig: positive, negative only}
\end{figure}

\subsection{Alignment of Discriminator with Human Preference } 
Since the discriminator assigns the token level continuous reward during PPO, it is important to analyze the discriminator’s alignment with human preferences. Directly obtaining token-level human labels is challenging due to the fine granularity required. Therefore, we examine the test set of the token-level preference obtained by GPT-4 using our token-level preference labeling framework, and the results show a token-level preference prediction accuracy of 81.4\% from our trained discriminator. This high accuracy indicates that the discriminator effectively mirrors token-level preferences, confirming the effectiveness of our approach in aligning with human preferences at the token level.

\subsection{The Effect of Positive and Negative Token-Level Rewards} 

In this section, we explore TLCR's performance under two conditions: excluding negative rewards (where we consider only $r_t \geq 0$ from Equation 7, denoted as TLCR$_\text{w/o negative}$) and excluding positive rewards (where we focus on instances where $r_t \leq 0$ in Equation 7, denoted as TLCR$_\text{w/o positive}$). Similar to \citet{secrets_of_rlhf1}, we monitor the PPO training process by using action space modeling metrics, such as perplexity, response length, and KL divergence between the policy model and the SFT model to show informative details of the training stability of each setting.

In Figure \ref{fig: positive, negative only}, under conditions excluding negative rewards, the model is observed to increase the generation length to increase the number of positive tokens, thereby inflating its reward. Conversely, sentence length decreases to minimize negative token generation when positive rewards are not used. It can be viewed as 'reward hacking',  resulting in highly elevated KL divergence and perplexity. This result shows the importance of considering both positive and negative rewards in training to prevent reward hacking and ensure accurate preference training. 

\begin{figure}[t!]
	\centering
    	\includegraphics[width=1.0\linewidth]{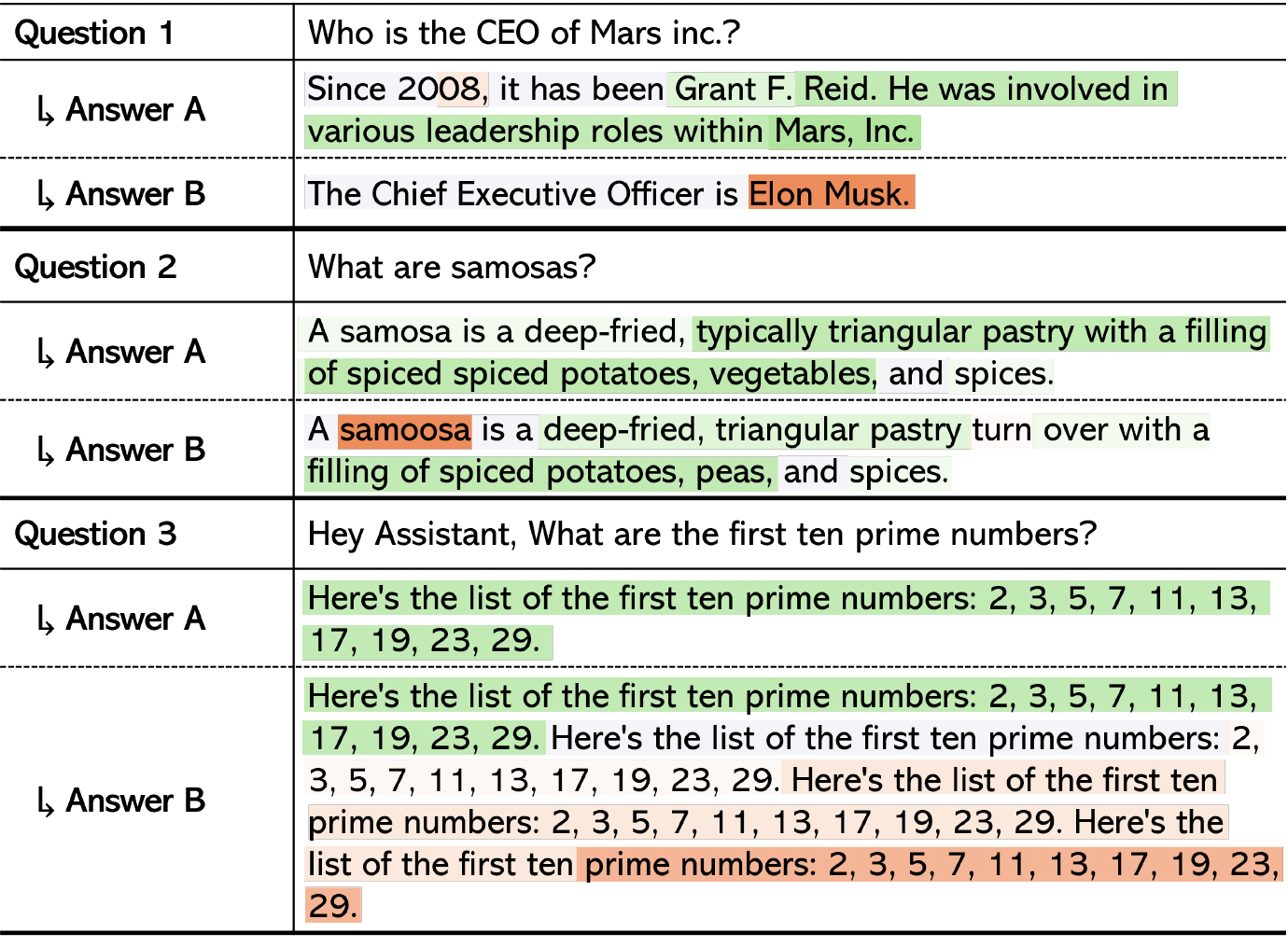}
	\caption{\textbf{Examples of Token-Level Preference Predictions from our trained Discriminator $D_\phi$.} Green represents positive preference, and red represents negative preference. The concentration of the color visualizes the intensity (i.e., scale) of the preference.} 
	\label{fig: qual}
\end{figure}

\subsection{Qualitative Results}

Figure \ref{fig: qual} illustrates the qualitative analysis of token-level preference predictions made by our trained Discriminator $D_\phi$. For the response A to Question 1, incorrect details like '08' from '2008' are identified as negative due to the misinformation. Conversely, the subsequent tokens are predicted as positive due to their accurate information. Similarly, in the case of response B, 'Elon Musk' receives a negative prediction, attributed to the inaccurate information.

Furthermore, in the response B to Question 2, the term 'samoosa' is marked as unfavorable, likely due to an apparent spelling error, contrasting with the favorable assessment of 'samosa' in response A, which is spelled correctly.

For Question 3, the response B exhibits hallucinated outputs, characterized by repetitive, non-terminating sentences. While the initial sentence in both response A and response B is awarded a favorable preference, the following hallucinated segments in response B are marked with a negative preference.

These examples highlight the value and usefulness of token-level preference in providing detailed feedback on sentence quality. As illustrated, a generated sequence can include a mixture of accurate and inaccurate information or even correct and helpful responses impaired by minor spelling errors in specific words. This fine-grained approach to feedback is crucial for distinguishing the subtle quality of language model outputs.


\section{Conclusions}
In this study, we introduced the Token-Level Continuous Reward (TLCR), a novel reward model aimed at providing detailed, token-based continuous rewards for Reinforcement Learning from Human Feedback (RLHF). By utilizing the prediction confidences from a discriminator to assign rewards at the token level, our approach offers a sophisticated reward system that adjusts to the varying degrees of relevance each token holds within its context. The superiority of TLCR over traditional sequence-level and token-level discrete reward mechanisms has been established through detailed experiments and analyses. 

\section{Limitations}

Current work uses a 7B parameter model, trained with a single dataset partitioned for three stages. Also, the current reward model training utilizes the offline static dataset, which might cause a distribution mismatch during policy model training. Moreover, the token-wise preference discriminator exhibits a bias due to the ambiguity of good human preference, implying room for improvement. Future works can include model-data scaling, iterative updating of the reward model and the policy model, and combining multi-objective preference (e.g., safety reward and helpfulness reward) learning for more fine-grained guidance.

\section{Broader Impact}
This paper presents a method to improve LLMs via RLHF, targeting fine-grained alignment with human preferences and values. We hope our work will help reduce potential risks caused by LLM generations.

\section{Acknowledgements}
This work was supported by Institute of Information \& communications Technology Planning \& Evaluation (IITP) grant funded by the Korea government (MSIT) (No. 2022-0-00951, Development of Uncertainty-Aware Agents Learning by Asking Questions), Institute of Information \& communications Technology Planning \& Evaluation (IITP) grant funded by the Korea government(MSIT) [RS-2021-II212068, Artificial Intelligence Innovation Hub (Seoul National University)] and Institute of Information \& communications Technology Planning \& Evaluation (IITP) grant funded by the Korea government (MSIT)(No. RS-2019-II190079, Artificial Intelligence Graduate School Program, Korea University).

\bibliography{custom}

\clearpage
\appendix
\onecolumn

\section{Prompt Used for Instructing the External Language Model}

\label{sec:appendix}
\begin{figure*}[h]
\begin{center}
\centerline{\includegraphics[width=0.8\linewidth]{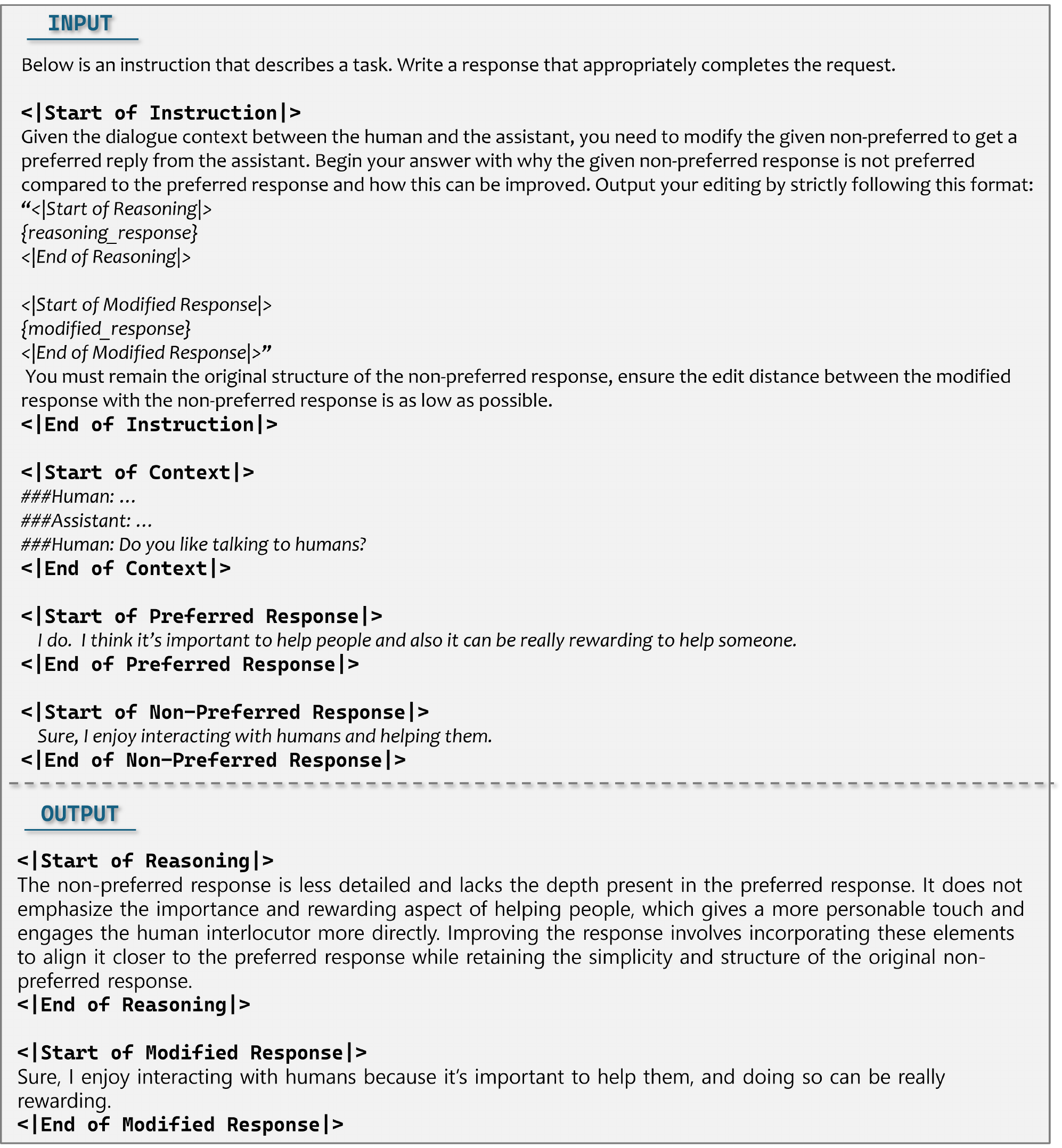}}
\caption{\textbf{Prompt used to instruct the external language model $\text{LM}_\text{ext}$ to modify the rejected sample.}}
\label{fig: Instruction example}
\end{center}
\end{figure*}

\section{Experimental Details}
\label{appendix: experimental_detail}
We use the Adam optimizer, setting $\beta_1 = 0.9$ and $\beta_2 = 0.95$. We set the maximum sequence length, including prompts and responses, to 512. Further details for each stages of the PPO process using the Deepspeed-Chat \cite{deepspeed} framework is detailed as follows:
\begin{itemize}
\item \textbf{Supervised Fine-Tuning (SFT):} We use a batch size of 16 per device, a learning rate of $10^{-5}$, and use a cosine learning rate schedule without warm-up. We train for 2 epochs.

\item \textbf{Reward Model Training (i.e., Discriminator $D_\phi$ training):} Our reward model training stage involves training the discriminator. Similar to the SFT training, we use the batch size of 16 per device, with a learning rate of $10^{-5}$, and use a cosine learning rate schedule without warm-up. A weight decay of 0.1 is used, and we train for 2 epochs.

\item \textbf{Reinforcement Learning with PPO:} For the RL stage, we use ZeRO-2, gradient checkpointing, and offload. The learning rate is set to $10^{-6}$, and we use a cosine learning rate schedule without warmup. We train for 1 epoch. The KL penalty coefficient is set to 0.1. We use ZeRO-3 for the reference model and the reward model, as they do not require training. The hybrid training engine is used.
\end{itemize}

\section{Generated Response Comparison}
\begin{table*}[hbt!]
	\centering
	\begin{tabular}[t]{l l}
		\Xhline{3\arrayrulewidth}
        \textcolor{cyan}{\bf{Instruction}}  & \makecell[l]{\textcolor{cyan}{What is some cool music from the 1920s?}} \\
		\Xhline{2\arrayrulewidth}
		\bf{SFT}  & \makecell[l]{1920s music?  I’m not sure I know what that means.  I’ll do my best to help.  Can you \\ tell me a little more about what you’re looking for?} \\ \hline
        \bf{PPO}  & \makecell[l]{1920s jazz music was hugely influential and popular, and some classic songs from \\ that era include “Some of These Days” by Ma Rainey, “Livery Stable Blues” by \\ Bessie Smith, “The Charleston” by Scott Joplin, and “I’m Crazy ‘Bout My Baby” \\ by Fats Waller. Other popular 1920s music genres included ragtime, big band, \\ and early blues.} \\ \hline
        \bf{FIGA}  & \makecell[l]{1920s music tends to be jazz and blues. If you like jazz, I recommend The Jazz \\ Singer by Al Jolson, or Hot Jazz from Louis Armstrong. If you’re interested in \\ blues, King Oliver's Creole Jazz Band is a great place to start, or any music by \\ Bessie Smith. These would all be great choices for a collection of cool 1920s music. \\ If you're looking for something more current, you might also enjoy artists \\ like Louis Armstrong, Duke Ellington, or Bix Biederbeck, who played a heavy role \\ in shaping music during that time period. You could also explore the music of the \\ Harlem Renaissance, which was a cultural movement that significantly influenced the \\ 1920s and beyond. Of course, this list is just a starting point, and there are many other \\ great artists and songs from the 1920s that are worth exploring. These are a few \\ suggestions to get you started on your journey into the music of this era, with plenty \\ of options to discover and enjoy.} \\ \hline
        \bf{TLC$\text{R}_{\text{fixed}}$}  & \makecell[l]{Here are 10 cool songs from the 1920s:\\
        -“Some of These Days” by Sophie Tucker\\
        -“When the Saints Go Marching In” by the Original Dixieland Jazz Band\\
        -“Tiger Rag” by Louis Armstrong\\
        -“Yes! We Have No Bananas” by Benny Goodman\\
        -“I Got Rhythm” by George Gershwin\\
        -“Ain’t Misbehavin’” by Fats Waller\\
        -“I’m Looking Over a Four-Leaf Clover” by the Mills Brothers\\
        -“Honeysuckle Rose” by Bix Beiderbecke\\
        -“The Charleston” by Glenn Miller\\
        -“I’m Confessin’ (That I Love You)” by the Ink Spots} \\ \hline
         \bf{TLCR (Ours)}  & \makecell[l]{1920s music can be cool in many ways. Here are some suggestions:\\
        - Jazz: This style of music featured improvisations and syncopated rhythms. \\ Notable jazz musicians from the 1920s include Louis Armstrong, Duke Ellington, \\ and Bix Beiderbecke.\\
        - Blues: The blues style originated in the 1920s, and featured simple melodies \\ and soulful vocals. Famous blues artists from this era include Ma Rainey, Bessie \\ Smith, and Charley Patton.\\
        - Gospel: This style of music began in the 1920s, combining elements of blues, \\ jazz, and spirituals. Notable gospel performers from the 1920s include the Campbell \\ Sisters, Mahalia Jackson, and Thomas A. Dorsey.\\    
        - Vaudeville: Vaudeville is a type of variety entertainment that became popular \\ during the 1920s. Its music featured fast-paced songs with lively rhythms, such as \\ ragtime, cakewalk, and novelty numbers. Famous vaudeville performers from the \\ 1920s include Bert Williams, Eddie Cantor, and Al Jolson.\\
        - Broadway: Broadway theater music became popular in the 1920s, featuring lavish \\ productions and elaborate dance routines. Notable Broadway composers} \\
        \hline
        \hline
    \Xhline{2\arrayrulewidth}
\end{tabular}
	\caption{\textbf{Example of the Generated Responses from Different Methods for a Given Instruction.}}
	\label{tab: generation example}
\end{table*}

\end{document}